\documentclass[10pt,twocolumn,letterpaper]{article}

\usepackage{cvpr}
\usepackage{times}
\usepackage{epsfig}
\usepackage{graphicx}
\usepackage{amsmath}
\usepackage{amssymb}

\usepackage[pagebackref=true,breaklinks=true,letterpaper=true,colorlinks,bookmarks=false]{hyperref}

\cvprfinalcopy 

\def\cvprPaperID{1508} 

\ifcvprfinal\fi
\begin{document}

\title{Towards Spectral Estimation from a Single RGB Image in the Wild}

\author{Berk Kaya\\
CVL, D-ITET, ETH Zurich\\
{\tt\small bekaya@student.ethz.ch}
\and
Yigit Baran Can\\
CVL, D-ITET, ETH Zurich\\
{\tt\small cany@student.ethz.ch}
\and
Radu Timofte\\
CVL, D-ITET, ETH Zurich\\
{\tt\small timofter@vision.ee.ethz.ch}
}

\maketitle

\begin{abstract}
In contrast to the current literature, we address the problem of estimating the spectrum from a single common trichromatic RGB image obtained under unconstrained settings (\eg unknown camera parameters, unknown scene radiance, unknown scene contents). For this we use a reference spectrum as provided by a hyperspectral image camera, and propose efficient deep learning solutions for sensitivity function estimation and spectral reconstruction from a single RGB image.
We further expand the concept of spectral reconstruction such that to work for RGB images taken in the wild and propose a solution based on a convolutional network conditioned on the estimated sensitivity function. Besides the proposed solutions, we study also generic and sensitivity specialized models and discuss their limitations. We achieve state-of-the-art competitive results on the standard example-based spectral reconstruction benchmarks: ICVL, CAVE, NUS and NTIRE. Moreover, our experiments show that, for the first time, accurate spectral estimation from a single RGB image in the wild is within our reach.\footnote{Our codes and models will be made publicly available upon acceptance of the paper.}
\end{abstract}

\section{Introduction}
\label{sec:introduction}

Unlike conventional RGB images, hyperspectral (HS) imagery captures more information from the electromagnetic spectrum and represents it using a higher number of spectral bands. This led hyperspectral image (HSI) processing to become a crucial field in many computer vision tasks such as object recognition, segmentation and anomaly detection~\cite{stein2002anomaly}. There are numerous applications  in remote sensing~\cite{bioucas2012hyperspectral, ozkan2018endnet}, medical diagnosis ~\cite{lu2014medical, martin2006development}, material detection~\cite{barbin2013non}, food inspection~\cite{lu1999hyperspectral} and agriculture~\cite{govender2007review} which make use of spectral data. However, capturing spectral data is difficult due to the limitations of the imaging technology. Commonly used imaging methods apply scanning in the spectral domain to acquire the full spectrum~\cite{garini2006spectral}. This acquisition process is time consuming and the equipment is relatively expensive. 
\begin{figure}[t!]
\centering
\includegraphics[width=.95\linewidth]{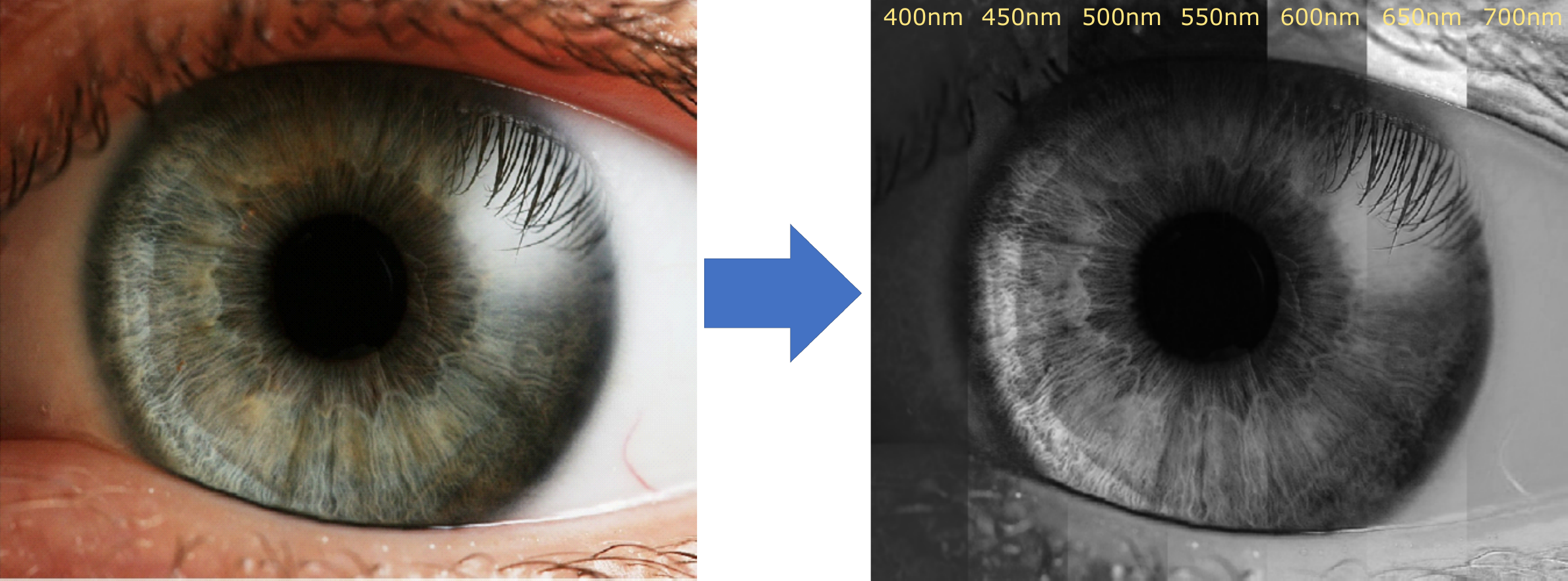}
\caption{Our proposed approach estimates spectral bands for an RGB image captured under unknown settings. Here, several bands are estimated for a random eye image in the wild.}
 \label{fig:teaser}
 \vspace{-0.55cm}
\end{figure}

One way to obtain spectral data is to infer the missing band information from the RGB image of the scene. This problem is referred to as \textbf{spectral reconstruction} or \textbf{spectral super-resolution}. It is an ill-posed problem since it tries to reconstruct a high number of bands (usually 31 uniformly sampled, 400nm to 700nm) from the RGB image which contains 3 bands. Nevertheless, it is possible to extract high level information from the RGB image which enables reconstruction of spectral bands. 
In recent years, spectral super-resolution has become an active field of research because it can be applied for the systems where hyperspectral cameras cannot be integrated. It also makes it possible to capture dynamic scenes due to the rapid acquisition property of RGB cameras. Critical for the spectral reconstruction accuracy of such methods are the knowledge of the camera parameters / sensitivities and image contents and physical properties, as well as the availability of training data under the form of paired spectral and RGB images.

\begin{figure*}[th!]
\centering
\includegraphics[width=0.95\textwidth]{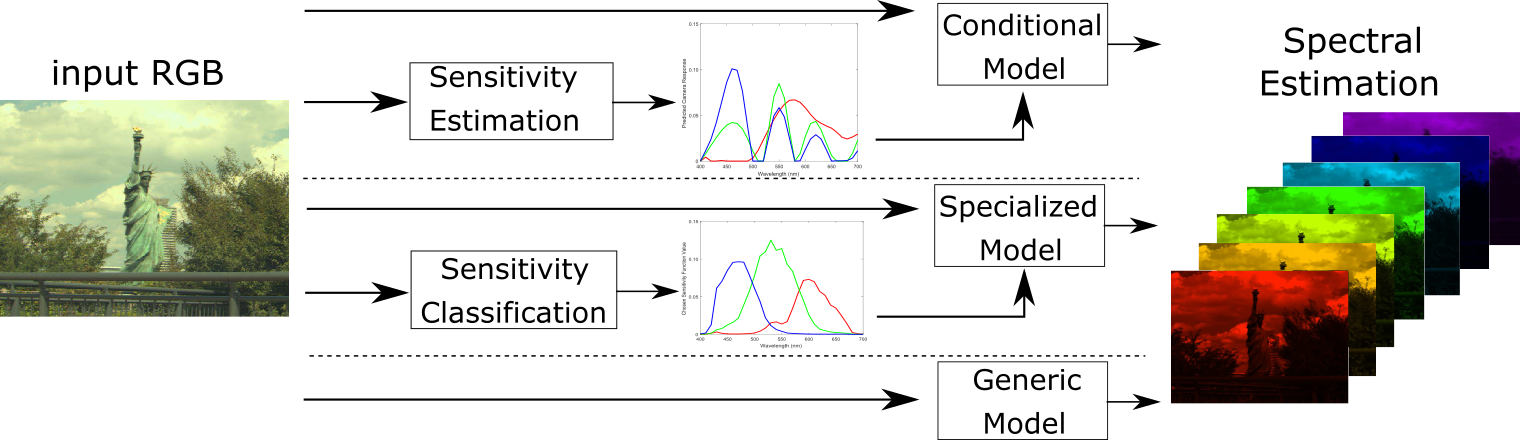}
\caption{The proposed conditional, specialized and generic pipelines for spectral estimation from a single RGB image input captured in the wild. (RGB image from~\cite{DBLP:conf/qomex/ImamogluOZDFKN18})}
\label{fig:pipelines}
\vspace{-0.2cm}
\end{figure*}

The are works~\cite{DBLP:conf/eccv/NguyenPB14,arad2016sparse,arad2018ntire,yan2018accurate,aeschbacher2017defense,DBLP:journals/corr/GallianiLMBS17} which employ example-based spectral reconstruction on RGB images. Most common approaches utilize sparse dictionary learning and deep learning methods. The main drawback of these methods is that they assume a CIE matching function to form a relation between the incoming spectrum and captured RGB values. However, this supposition is not true since cameras generally have different spectral sensitivity functions some of which are not similar to CIE~\cite{arad2017filter}. This causes the implementations to produce inaccurate results~\cite{nie2018deeply}.  Therefore, the sensitivity function of the camera must be used together with RGB data to perform spectral reconstruction. However, the sensitivity function in not always known by the user in many applications, especially in the case where the source of the captured RGB image is unknown.

In this paper we aim at spectral estimation from a single RGB image captured under unknown settings, in the wild. For this purpose, we propose a framework consisting of an estimator CNN model that estimates the sensitivity function given an RGB image and a reconstruction model that takes RGB image as input and conditioned on the estimated sensitivity, produces spectral estimation of the scene. 

In this framework, we also propose an efficient CNN architecture to be used as the reconstruction model. Apart from demonstrating the performance of the proposed framework, we also validate the proposed efficient reconstruction model on four standard benchmarks for spectral reconstruction from RGB images. 

We also consider controlled cases where the required sensitivity is a member of a finite set of functions. For such cases, we experimented with a CNN-based classifier along with our estimator model to provide a comparison. 

The main contributions of this paper can be summarized as follows:
\begin{enumerate}
\item We propose an efficient deep learned solution to example-based spectral reconstruction from a single RGB image and achieve state-of-the-art results on standard benchmarks: ICVL, CAVE, NUS, NTIRE.
\vspace{-0.2cm}
\item We introduce a deep learning solution for estimating the sensitivity function from a single RGB image input having a reference spectrum.
\vspace{-0.2cm}
\item We expand the concept of spectral reconstruction from a single RGB image for in the wild settings. The sensitivity function is first estimated and then used to perform spectral estimation.
\vspace{-0.2cm}
\item We study the proposed method along generic and sensitivity specialized models.
\end{enumerate}

\section{Related Work}
\label{sec:related_work}

The spectral responses of digital cameras have significant effects on image formation process. For this reason, these response functions are extensively studied in the literature including the ways to represent them more efficiently. There are several studies which try to regress the sensitivity function using measurement setups. Nevertheless, the estimated sensitivity information is never used as a prior to the spectral reconstruction problem before. 

\textbf{Spectral sensitivity functions.} These functions generally exhibit similar characteristics due to the semiconductor sensor designs adapted to human vision system. 
In common estimation procedures, monochromatic light sources or narrow band filters are used to illuminate a target. Then, the response of the camera is recorded for each channel considering the wavelength target is exposed to \cite{darrodi2015reference}. This procedure is costly and time consuming to determine the response of a single camera. Therefore, the methods which builds a relation between scene radiance and RGB recordings are more preferred. 

\textbf{Sensitivity estimation with known radiance and RGB images.} If the scene radiance and RGB values are provided, the spectral response of a camera can be estimated using statistical methods~\cite{finlayson2016rank}. Such a recovery is not possible with standard least square solution since the rank of the problem does not allow us to interpret realistic sensitivity responses. Tikhonov regularization is used to estimate the mapping between RGB and spectral data because of the fact that it allows us to form a realistic solution. This is achieved by adding a regularization term which calculates second derivative vectors, imposing smoothness condition for the sensitivities~\cite{darrodi2015reference}. Li~\etal~\cite{li2018efficient} proposed to learn an optimized training set of RGB-hyperspectral pairs and use radial basis function interpolation to infer spectrum of a given image while assuming the spectral power distributions of illumination is known.


\textbf{Sensitivity estimation without spectral data.} Methods without the usage of spectral data directly stem from the hypothesis that the radiance of some materials can be estimated beforehand. In \cite{huynh2014recovery}, a color chart is exposed to spectrally smooth illumination and a constrained minimization problem is solved to calculate sensitivities. Similarly, in \cite{jiang2013space}, a similar problem is solved by representing spectral responses using Principal Component Analysis(PCA).  These works assume that the illumination type is same and use a color chart whose reflectance is known. Another work \cite{han2012camera}, uses a flag made of fluorescence to have a prior information regarding the radiance and calculates sensitivity accordingly. In \cite{kawakami2013camera}, single image estimation is applied only for sky images, whose radiance is assumed to be known by the user of the method. Although there are various approaches existing for this analysis, \textit{multispectral data is used along with the RGB image or a flag object is used to impose extra information}  to the optimization method. In this manner, our method differs from any other algorithm since \textit{no prior information is given to the network along with the training data.}

\textbf{Spectral super-resolution.} There are huge number of super-resolution methods applied for spatial domain but only a limited number of works have been published for spectral domain. This is due to the fact that the problem is heavily underconstrained \ie one has to predict more than 30 channels only from 3 channel values provided. However, it is still possible to upsample three channels to more since most of the hyperspectral bands are highly correlated.

In the method proposed by Arad~\etal~\cite{arad2016sparse}, they introduce a sparse dictionary from high resolution HS data and use this dictionary with Orthogonal Matching Pursuit (OMP) in order to perform sparse spectral reconstruction. However, most of the state of the art methods stem from Convolutional Neural Networks (CNNs). Recently, Galliani~\etal~\cite{DBLP:journals/corr/GallianiLMBS17} proposed a variant of a \textit{Tiramisu}~\cite{jegou2017one} network architecture which is generally used for semantic segmentation tasks. Aeschbacher~\etal~\cite{aeschbacher2017defense} proposed a new spectral upsampling method based on the A+ super-resolution method~\cite{timofte2014a+}.
Interestingly, Oh~\etal~\cite{Oh_2016_CVPR} proposed to use several consumer-level digital cameras with known spectral sensitivities to optimize hyperspectral imaging. 

Spectral reconstruction from a single RGB image state-of-the-art results are obtained by CNNs. Various types of deep CNN architectures are presented in~\cite{arad2018ntire}. In this paper, we base our experiments on a novel moderately deep architecture. This network is especially suitable for us because it requires much less memory, computational resources and time to run compared to other recent CNN based state-of-the-art methods. This allows us to build more complex and memory intensive systems to work together with this network. We also make modifications to this network to adapt it for different sensitivities other than CIE matching function.

\begin{figure}[t]
\centering
\includegraphics[width=0.4\textwidth]{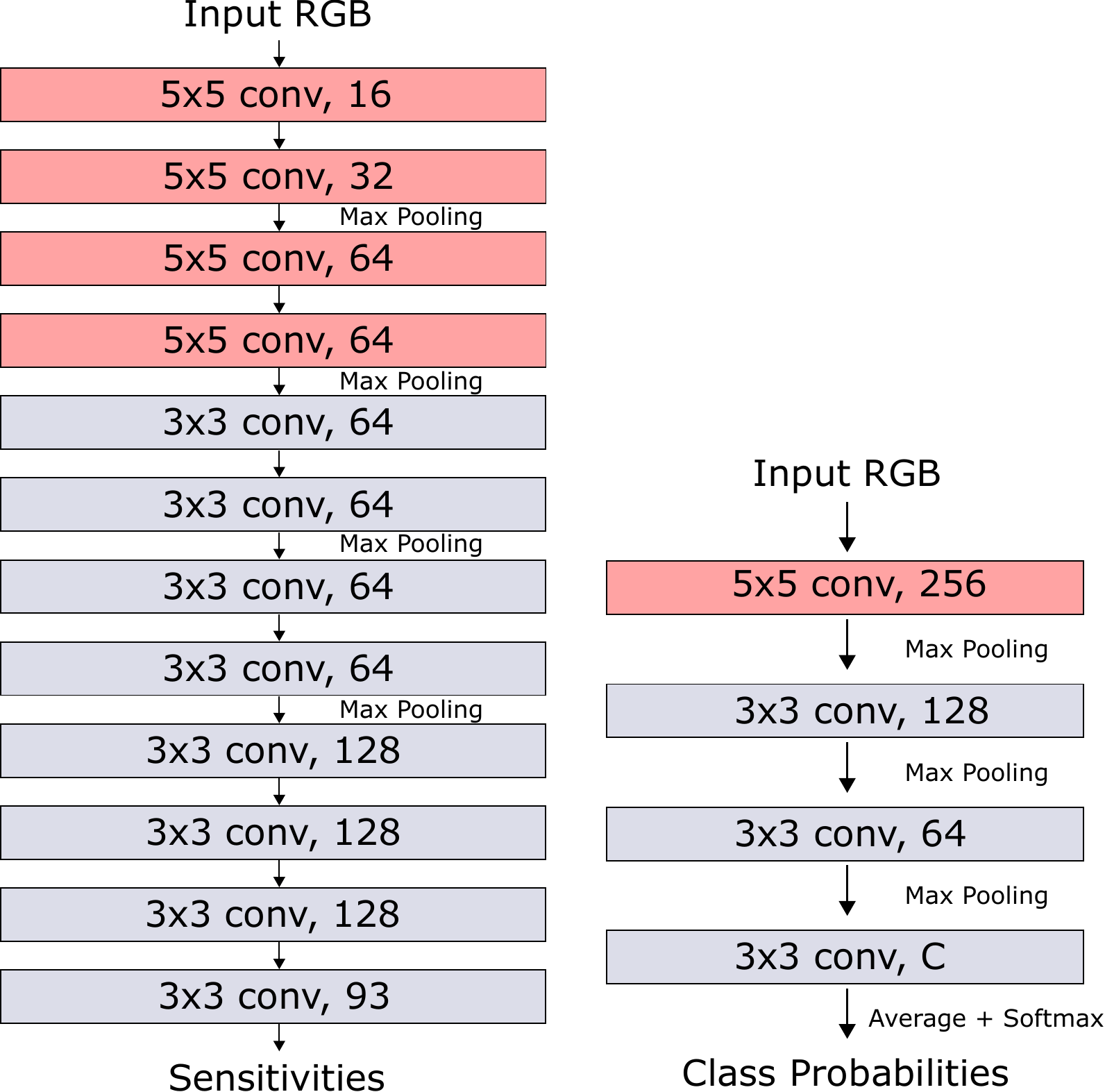}
\caption{Architecture of the proposed sensitivity estimation and sensitivity function classification networks.}
\label{fig:kaya_network}
\vspace{-0.2cm}
\end{figure}

\section{Proposed Methods}
\label{sec:proposed_methods}

In this section we introduce our models for sensitivity function estimation and classification, and spectral reconstruction for conditional, specialized, and generic settings.

\noindent\textbf{Image Formation. }
The channel readings of RGB sensors are integrations over the visible spectrum $V$.
\begin{equation}
	I_c(u) =  \int_{V}^{} S_c(\lambda)L(u,\lambda)d\lambda,
\end{equation}
where $L(u,\lambda)$ defines the spectral radiance corresponding to point $u$ and $S_c(\lambda)$ is the spectral sensitivity of the channel $c\in\{R,G,B\}$.
In discrete settings we have 
\begin{equation}
	I_c(u) = \sum_{n}^{} S_c(\lambda _n)L(u,\lambda _n),
\label{eq:sensitivity_discrete}
\end{equation}
where $\lambda_n$'s are sampled wavelengths~\cite{nie2018deeply} (usually 31). 
In our experiments we use the sensitivity functions and HS data to generate RGB images.

\subsection{Sensitivity Estimation from an RGB Image}
\label{ssc:sensitivity_estimation}

We propose a deep learning approach for estimating the parametrization of the sensitivity functions from a single RGB image input.

\noindent\textbf{Parametrization.} 
The estimation of the sensitivities is challenging because of the under-constrained nature of the problem \ie distinct sensitivity functions may result in similar RGB mappings.
However, we found out that its accuracy is not crucial for the task of spectral reconstruction as it is the reconstruction network's job to weigh in this additional information and the RGB image and infer the correct mapping.
We represent the sensitivity function in its discrete form~\eqref{eq:sensitivity_discrete}, while in literature there are other parameterizations with fewer parameters (such as in~\cite{zhao2009estimating}).

\noindent\textbf{Sensitivity estimator network.} 
The proposed fully convolutional network is given in Figure~\ref{fig:kaya_network}.
It consists of 12 convolutional and 4 maxpooling layers. The layers use ReLU activation and the output layer has
$3\times d$ feature mappings, where $d$ is the number of channels in HS data.
In forward propagation, the output block is averaged out in spatial axis to produce a single vector containing red, green and blue responses in combined form.
This vector is then shaped to its matrix form $S\in \mathbb{R}^{d\times 3}$.

\noindent\textbf{Sensitivity estimation loss.} 
Distinct sensitivity functions can lead to approximately same camera responses (one example is shown in Fig.~\ref{fig:rgb_recon}), an ambiguity very difficult to solve.
Therefore, we are not interested in the accurate estimation of the sensitivity function parameters and adjust our loss function in a way which prioritize the
difference between our input image and the image reconstructed by the estimated sensitivity:
\begin{equation}
	L_i = \frac{1}{n} \left\| HS - H\hat{S} \right\|_\textit{F}^2 =  \frac{1}{n} \left\| I - H\hat{S} \right\|_\textit{F}^2
\end{equation}
where $H\in \mathbb{R}^{n\times d}$ is the hyperspectral image, $I\in \mathbb{R}^{n\times 3}$ is the input RGB and 
$\hat{S}\in \mathbb{R}^{d\times 3}$ is the estimated sensitivity function. 
In the training process, we calculate the squared Frobenius norm of the difference between input and reconstructed RGB images by using the HS data.
We also introduce a mean squared loss function related to the labels (sensitivities):
 \begin{equation}
	L_l = \left\| S - \hat{S} \right\|_\textit{F}^2 
\end{equation}

\begin{figure}[t]
\centering
\includegraphics[width=0.4\textwidth]{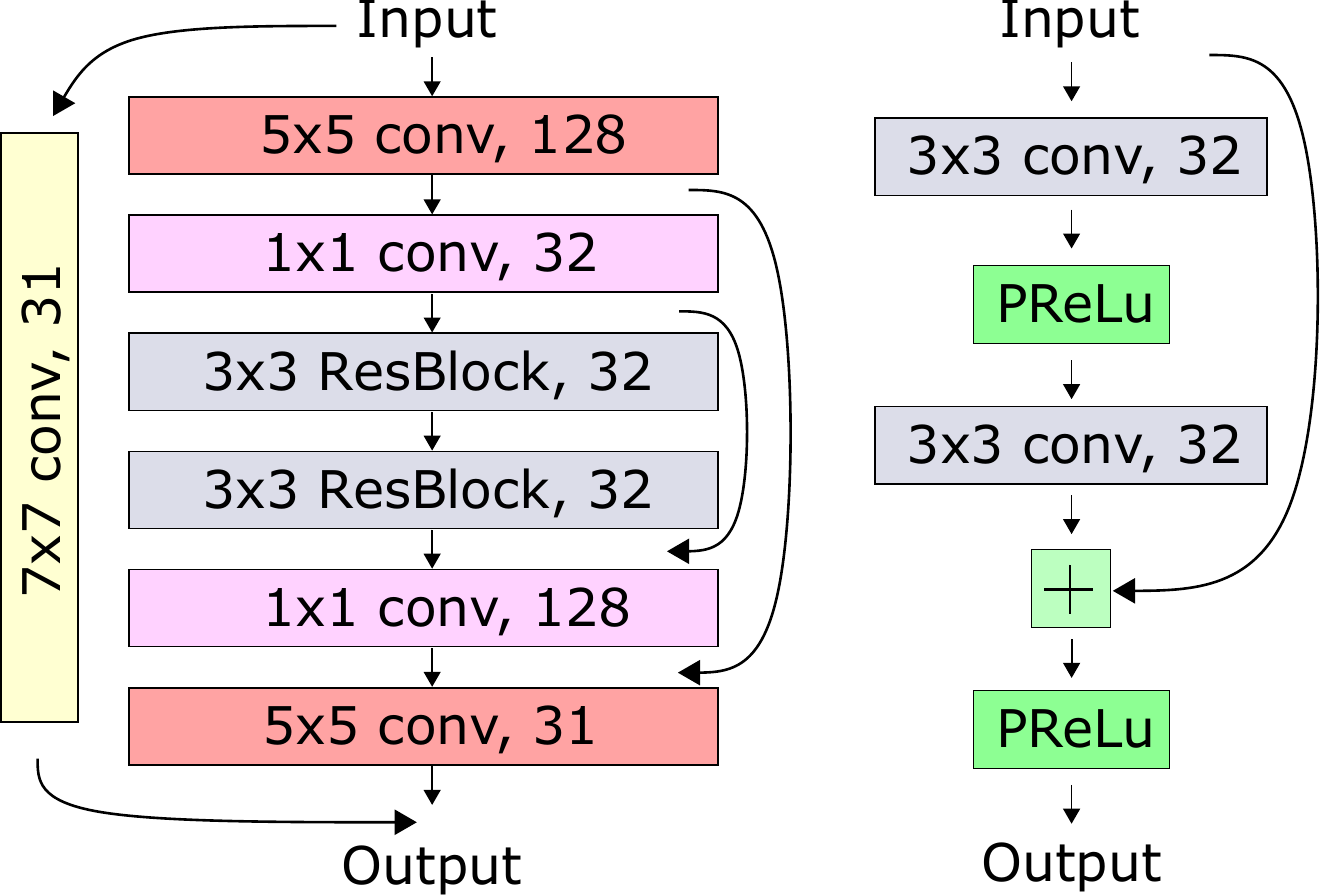}
\caption{Architecture of the spectral reconstruction network (left) and the structure of the residual block (right).}
\label{fig:baran_network}
\vspace{-0.2cm}
\end{figure}

We also regularize the sensitivity functions by calculating their second derivative vectors.
This regularization creates a smoothness effect and prevents aberrations from real world sensor responses.
In order to enforce this regularization, we introduce a 2\textsuperscript{nd} derivative operator $T\in \mathbb{R}^{(d-2)\times d}$
and the smoothness regularization loss is calculated as
  \begin{equation}
	L_s= \left\| TS \right\|_\textit{F}^2
\end{equation}

The total loss function of the network is the following:
\begin{equation}
	L= \lambda_1 L_i + \lambda_2 L_l + \lambda_3 L_s 
\end{equation}

 
\subsection{Sensitivity Classification from an RGB Image}
\label{ssc:sensitivity_classification}

In controlled settings, where the the cardinality of the set of sensitivity functions is limited, it is possible to implement a classifier model.
The aim of such a model is to predict which function is used to form that image so that spectral reconstruction can be performed.
In this paper, we propose a simple classifier network as shown in Figure~\ref{fig:kaya_network}.
The features obtained by the last convolutional layer are averaged and turned into probabilities by using softmax function.
Then, cross entropy loss between probability maps and labels are used to train the network.

\subsection{Spectral Reconstruction from an RGB Image}
\label{ssc:spectral_reconstruction}

For example-based spectral reconstruction from an RGB image we propose a moderately deep network designed to avoid overfitting to the training data.
The architecture can be analyzed in two parts. The core section consists of several convolutional layers, two skip connections, and two residual blocks as in Figure~\ref{fig:baran_network}.
There is also another branch which behaves as a skip connection and forms a basic mapping to the output.
The summation of these two branches builds a spectral reconstruction of the image.  $L_2$ loss function is used to minimize the reconstruction error.
However, the solution we pursue must work for several sensitivity functions which may be a set of continuous mappings.
Therefore, we derive several modes of operation for the network.

\begin{figure}[t]
\centering
\includegraphics[width=0.85\linewidth]{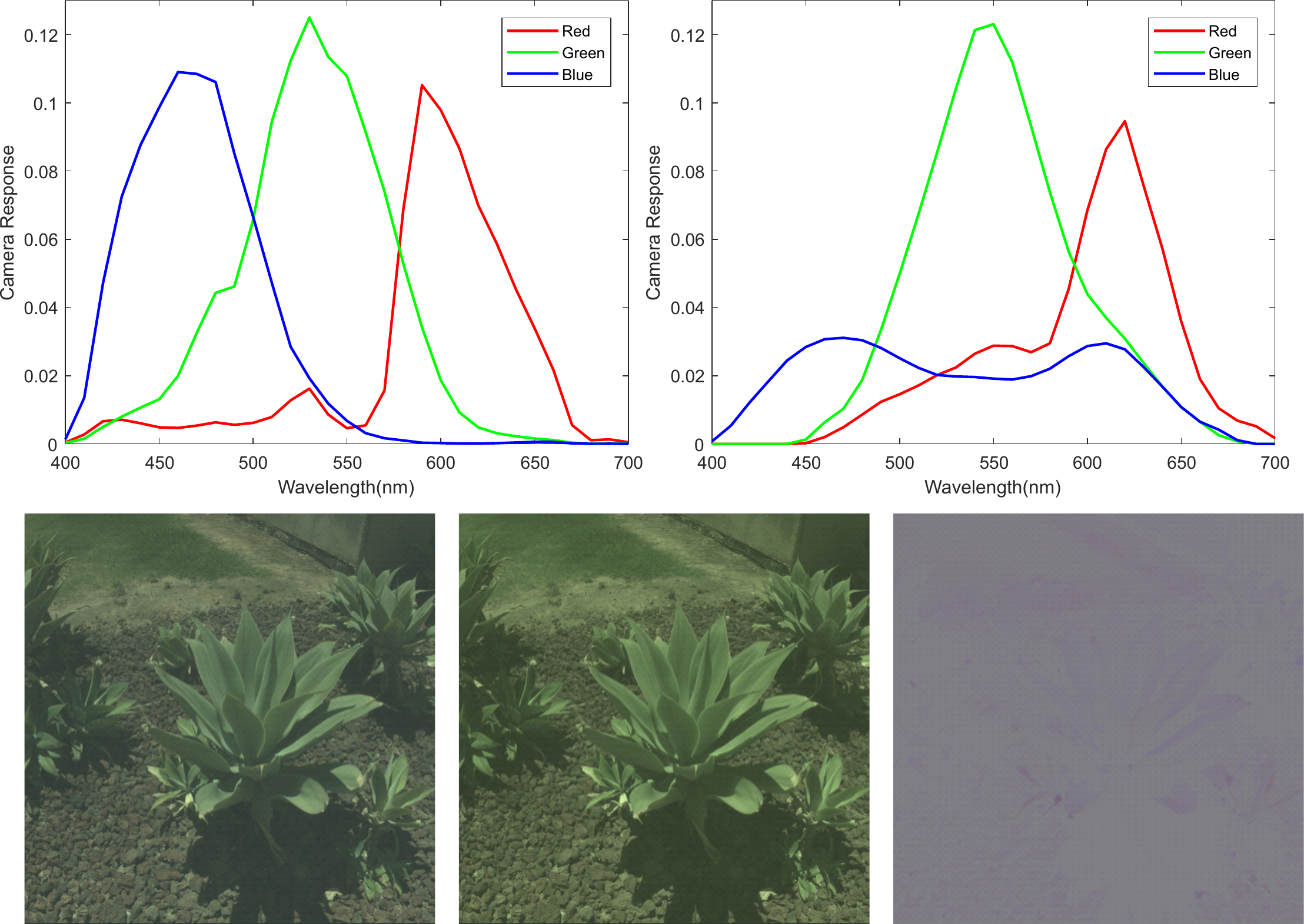}
\caption{Example of two sensitivity functions (top), corresponding RGB reconstructions from the spectral data and their difference (bottom). Gamma correction was applied for visualization.}
\label{fig:rgb_recon}
\vspace{-0.2cm}
\end{figure}

\noindent\textbf{Generic model.} 
For the generic model we train the network with images generated by different sensitivity functions without providing any
additional information about the sensitivity function. The model is expected to learn a mapping from RGB to HS
and adapt this mapping according to the input. 

\noindent\textbf{Conditional model.} 
Unlike the generic model, the conditional model gets the sensitivity information along with the RGB image input.
The sensitivity of three channels in single vector form are added as extra channels to the RGB image input.
In other words, each image pixel contains three sensor readings and the sensitivity function.
Since in the wild the sensitivity is not known, it has to be estimated by our estimator model beforehand.

\noindent\textbf{Specialized models.}
Another way of using the spectral reconstruction architecture is forming specialized models for a limited number of sensitivity functions.
Models can be trained to create a mapping for each function separately.
Still, in the wild, a model selection must be made. This is achieved by using the classification network described.

\section{Experiments and Results}
\label{sec:experiments}

In this section we first describe the experimental setup and discuss the results. For more details and (visual) results we invite the reader to check the supplementary material.

\begin{table}[t]
\small
\caption{\small{Average RMSE, PSNR, MRAE and SSIM  of the images reconstructed with the estimated sensitivity for different settings. Images are normalized to [0,1].}}
\label{tab:continuous}
\centering
\vspace{0.1cm}
{
\resizebox{\linewidth}{!}{
\begin{tabular}{l|c|c|c|c|c|c|c|c}    

    &\multicolumn{4}{c|}{\textbf{ICVL Dataset}}      &\multicolumn{4}{c}{\textbf{CAVE Dataset}} \\
    \hline
    
    \textbf{Training Set} &\multicolumn{2}{c}{Continuous}      &\multicolumn{2}{|c|}{Discrete} &\multicolumn{2}{c}{Continuous}      &\multicolumn{2}{|c}{Discrete}\\
    \hline
    
  \textbf{Testing Set}  & Cont & Disc &  Cont &  Disc &  Cont &  Disc&  Cont &  Disc\\

    \hline
    \hline
    RMSE $(\times10^{-2})$ & $   2.28   $ &$  2.82    $  & $   5.10   $ &$    1.47  $ &$   3.16    $ & $ 3.16      $ &$   3.80    $ & $ 2.92      $ \\
    \hline
    PSNR  & $ 33.35     $ &$  33.61    $ & $ 26.40     $ &$  39.34    $ &$ 28.83       $ & $    28.43   $&$ 28.78       $ & $    34.16   $\\
     \hline
    MRAE  & $  0.08    $ &$  0.13    $& $   0.16   $ &$  0.05  $ &$    0.21    $ & $   0.38    $&$   0.24     $ & $    0.16  $\\
     \hline
    SSIM  & $   0.98   $ &$  0.97    $ & $   0.97   $ &$   0.99   $&$   0.95     $ & $  0.92     $&$   0.94     $ & $  0.97    $\\
         \hline

\end{tabular}
}
}
\vspace{-0.3cm}
\end{table}


\subsection{Datasets}
\label{ssc:datasets}

In order to evaluate the performance of the proposed methods, we use four hyperspectral datasets commonly used in the literature:
ICVL~\cite{arad2016sparse}, CAVE~\cite{yasuma2010generalized}, NUS~\cite{nguyen2014training} and NTIRE~\cite{arad2018ntire} under their default benchmarking settings.
The default settings of these datasets assume HS images with 31 wavelengths uniformly distributed between 400nm and 700nm of the visual spectrum
and corresponding RGB images generated using specific response functions.
We refer the reader to the original works and our suppl. material for more details.

\begin{figure}[b]
\vspace{-0.5cm}
\centering
\includegraphics[width=0.45\textwidth]{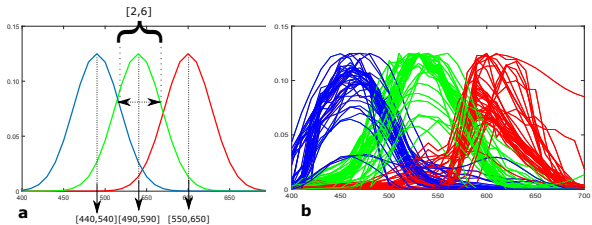}
\caption{\textbf{a)} Continuous set is created by sampling from the \textbf{uniform} distributions for means and standard deviations
of each Gaussian where a sensitivity function is assumed to be combination of Gaussians, \textbf{b)} all 40 functions of discrete set are shown.}
\label{fig:continuous_vs_discrete}
\end{figure}

\begin{figure*}[t]
\centering
\includegraphics[width=0.95\textwidth]{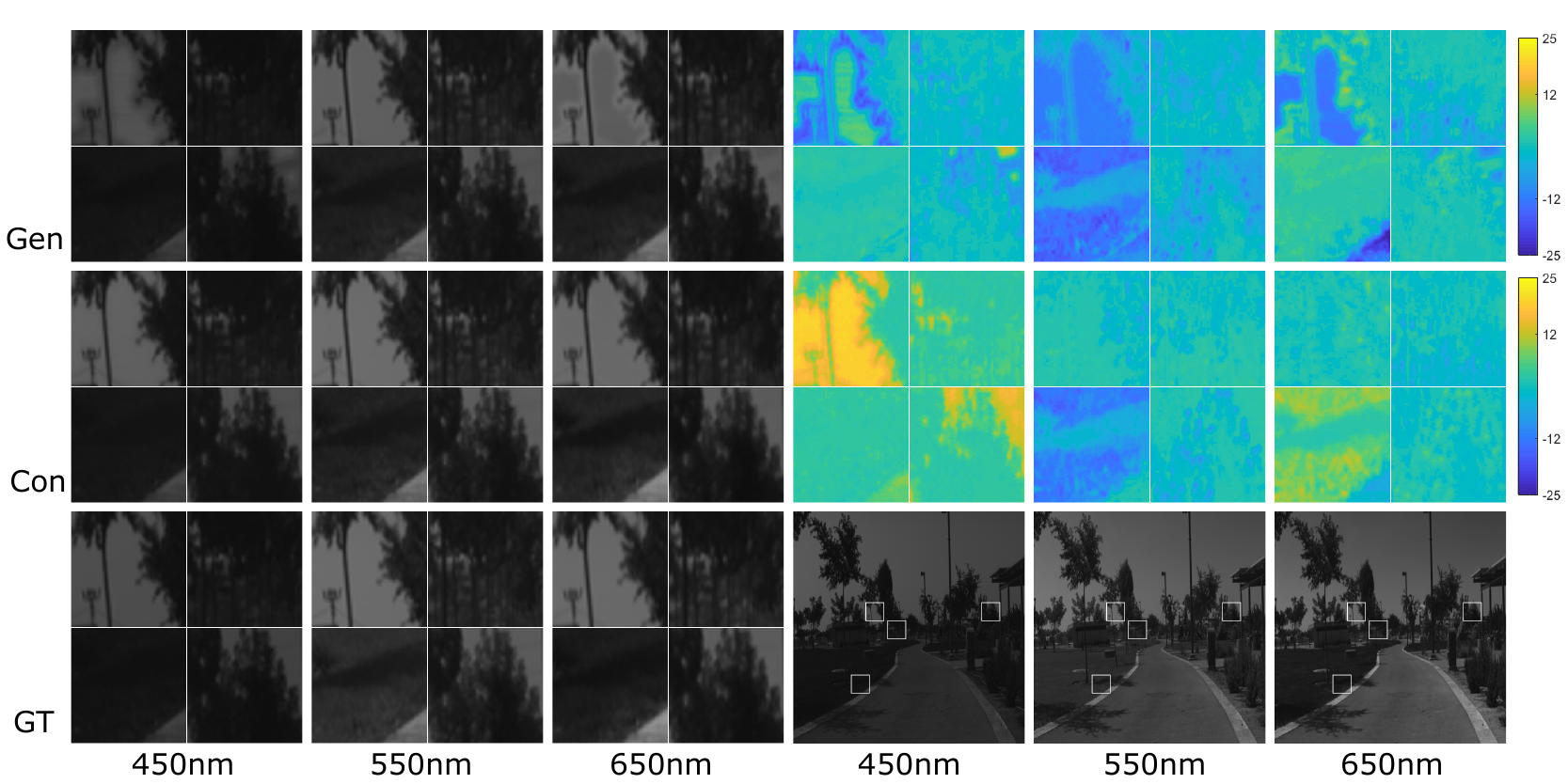}
\caption{Visual comparison between generic and conditional model spectral reconstructions and the groundtruth for an ICVL image. This is a case where the generic model performs better than the conditional model for bands below 500nm.}
\label{fig:ICVL_visual}
\end{figure*}

 \begin{figure}[t]
\centering
\includegraphics[width=0.4\textwidth]{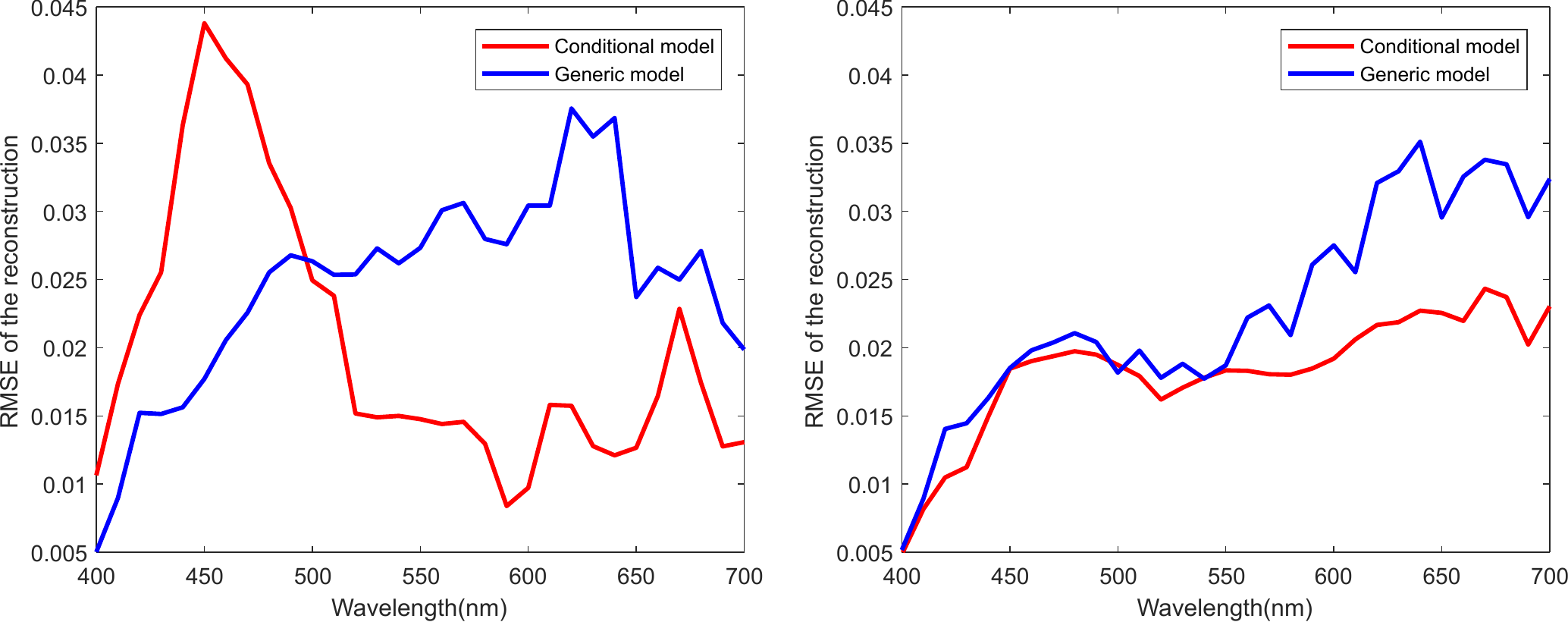}
\caption{RMSE corresponding to each band for the images compared in Figure~\ref{fig:ICVL_visual} (left) and average RMSE corresponding to each band for the ICVL test set (right).}
\label{fig:ICVL_visual_rmse_vs_band}
\vspace{-0.1cm}
\end{figure}


\subsection{Synthetic Data and Sensitivity Functions}
\label{ssc:synthetic}

We want to perform sensitivity estimation and spectral estimation from RGB images in the wild which is applicable to all camera types.
For this purpose use the ICVL data and its spectrum as reference.
We augment the ICVL data with data generated using sensitivity functions sampled from a continuous set or from a discrete set corresponding to real cameras.
These sets are described next and illustrated in Fig.~\ref{fig:continuous_vs_discrete}.

\noindent\textbf{Continuous Set.} 
Using only a limited number of mappings is not only impractical for application, but also prone to overfitting.
Therefore, the span of all possible sensitivity responses must be covered in training.
For this reason, we introduce a Gaussian Mixture Model (GMM) to randomly generate sensitivities.
A randomly generated response is modeled with the following expression for 31 channels.
\begin{equation}
S_c(x) = \alpha \sum_{j=1}^{k} \pi_j exp(\frac{x-\mu_j}{h_j^2})  \quad   s.t. \quad \sum_{j=1}^{k} \pi_j=1
\end{equation}
Here, $x\in\{1,2,...,31\}$ is the channel index and $Q_c(x)$ is the mapping for color $c\in\{R,G,B\}$. $\pi_j\in[0,1]$ are
the mixture ratios and we use $k\leq5 $  to limit the number of components in the mixture. To differentiate the spaces of colors,
we limit $\mu_j\in[16,26]$ for red,  $\mu_j\in[10,20]$ for green and $\mu_j\in[5,15]$ for blue channels. 
$h_j\in[2,6]$ is the parameter determining the variance and $\alpha$ is the scaling factor.
We also use $\alpha=\frac{1}{8}$ such that our RGB images are mapped to the same range with the HS data. 

\noindent\textbf{Discrete Set.} 
Apart from the randomly generated responses, we aim to evaluate our models for real world data.
Therefore, we use the combined dataset of two sensitivity recordings by applying same scaling factor.
The first dataset, provided by Kawakami~\etal~\cite{kawakami2013camera}, has measurement recordings for 12 camera brands between 400-700 nm with 4 nm intervals.
We used linear interpolation to get 31 channels with 10 nm increments.
The second dataset, provided by Jiang~\etal~\cite{jiang2013space}, consists of 28 camera measurements.
It covers the range of 400-720 nm with 10 nm intervals.
By using all these measurements together, we obtain a set consisting of 40 sensitivity responses.

\subsection{Implementation Details}
\label{ssc:implementation_details}

The methods covered in the paper include the utilization of three network architectures with different settings. The training procedure and hyperparameters can be found in the suppl. material.
The whole dataset is scaled such that the maximum radiance value is equal to one. For sensitivity estimation architecture,
the input RGB images are rendered from ICVL hyperspectral data using either continuous or discrete sensitivity sets depending on the mode of operation.
\begin{figure}[h]
\centering
\includegraphics[width=\linewidth]{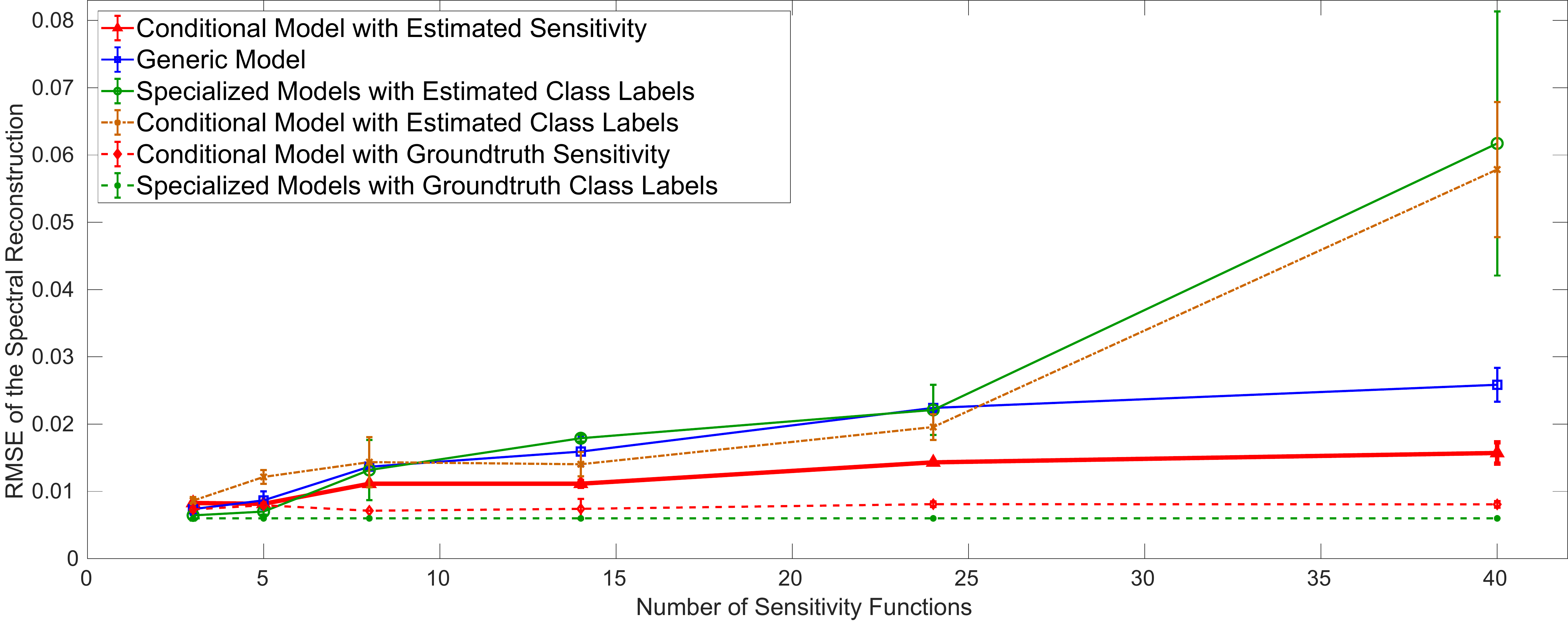}
\caption{RMSE of the reconstructions with respect to number of sensitivities allowed for different model settings.}
\label{fig:std_graph}
\vspace{-0.2cm}
\end{figure}
Spectral reconstruction model is trained independent of estimation or classification models. Same settings are used for generic, conditional and specialized models.
The validation errors are calculated throughout the training. The parameters which result in lowest validation errors are selected to be used in the testing setups (suppl. material).

\begin{figure}[t]
\centering
\includegraphics[width=0.8\linewidth]{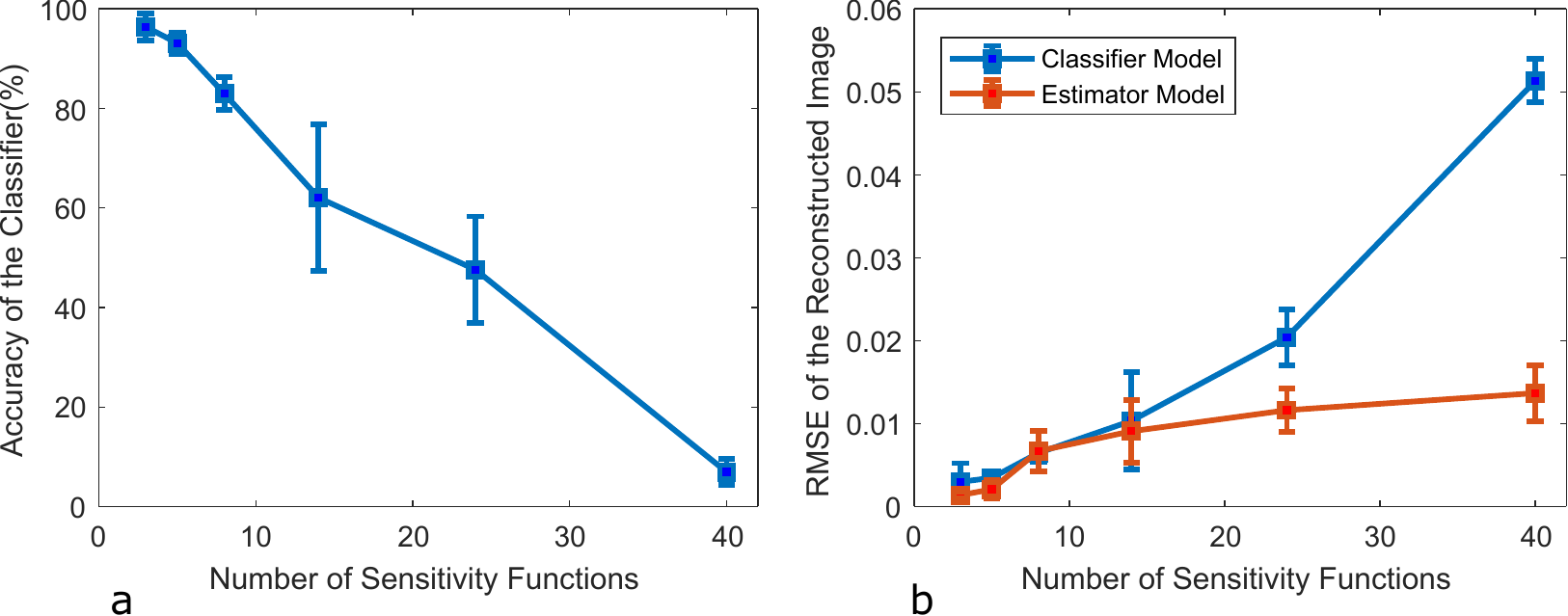}
\caption{a) Accuracy of the classifier on ICVL test set for different number of sensitivity functions;
b) Resulting RMSE of the reconstructed RGB image for classifier and estimator models.}
\label{fig:classification}
\vspace{-0.5cm}
\end{figure}

\begin{figure*}[t]
\centering
\includegraphics[width=0.95\textwidth]{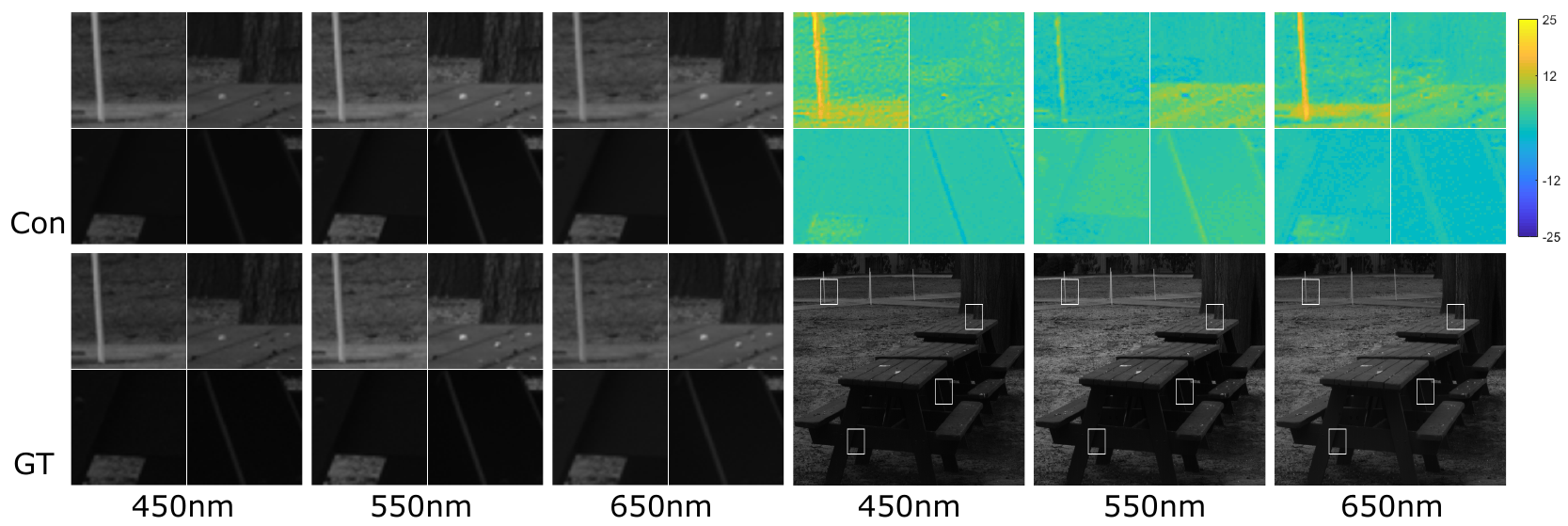}
\caption{Visual comparison on Harvard dataset~\cite{chakrabarti2011statistics} between conditional model spectral reconstructions (networks trained on ICVL dataset) and corresponding spectral images from Harvard dataset. More in the wild examples are shown in the supplementary material. }
\label{fig:harvard_visual}
\vspace{-0.1cm}
\end{figure*}

\subsection{Experimental Results}
\label{ssc:experimental_results}

\noindent\textbf{Loss functions and metrics.}
As stated previously, the performance of the estimator model is assessed by the difference between the input RGB and the RGB formed with the output sensitivity.
Figure~\ref{fig:rgb_recon} demonstrates sample sensitivity responses and RGB images generated using them.
Although the curves appear to be completely different, the corresponding images are almost the same.
Root Mean Squared Error (RMSE), Peak Signal-to-Noise Ratio (PSNR), Mean Relative Absolute Error (MRAE), and Structural Similarity (SSIM)
values are calculated for the these two images as evaluation metrics.
Table~\ref{tab:continuous} demonstrates the performance of our estimation model on ICVL and CAVE data for continuous and discrete sensitivity sets.
Although the continuous set of sensitivities are used to achieve a better generalization, the discrete set can also be used to train the estimator network.
As expected, each model performs better on the set which is used in training.

\begin{table*}[th!]
\caption{Quantitative comparison on ICVL, CAVE, and NUS benchmarks.
We report relative RMSE (rRMSE) and RMSE scores as in~\cite{arad2016sparse,aeschbacher2017defense} and
rRMSE$_G$ and RMSE$_G$ as defined in~\cite{DBLP:journals/corr/GallianiLMBS17} 
also after conversion to \textit{uint} precision. Images are normalized to [0, 255].}
\label{tab:quantitative_comparison_ICVL_CAVE_NUS}
\centering
\vspace{0.1cm}
\resizebox{\linewidth}{!}
{
\begin{tabular}{l|ccccc|ccccc|cccccc}
&\multicolumn{5}{c}{\textbf{ICVL dataset~\cite{arad2016sparse}}}&\multicolumn{5}{c}{\textbf{CAVE dataset~\cite{yasuma2010generalized}}}&\multicolumn{6}{c}{\textbf{NUS dataset~\cite{nguyen2014training}}} \\
 & Galliani\cite{DBLP:journals/corr/GallianiLMBS17}
 & Arad\cite{aeschbacher2017defense}
 & A+\cite{aeschbacher2017defense} &
 \textbf{ours}& \textbf{ours+E}
 & Galliani\cite{DBLP:journals/corr/GallianiLMBS17} 
 &Arad\cite{aeschbacher2017defense} 
 & A+\cite{aeschbacher2017defense} 
 & 
 \textbf{ours} 
 & \textbf{ours+E}
 & Nguyen 
 & Galliani\cite{DBLP:journals/corr/GallianiLMBS17} 
 & Arad\cite{aeschbacher2017defense} 
 & A+\cite{aeschbacher2017defense} 
 & \textbf{ours} & \textbf{ours+E} \\
 
\hline
\small{rRMSE} & -& 0.0507& 0.0344& 0.0168& \textbf{0.0166} &-  &0.4998 & 0.4265 & 0.4697& \textbf{0.178} & 0.2145 & -& 0.1904& \textbf{0.1420}&0.1524&0.1471  \\

\small{rRMSE$_G$} & -& 0.0873& 0.0584& 0.0401& \textbf{0.0399}&- & 0.7755& 0.3034 &0.246 &\textbf{0.239}& 0.3026&- & 0.3633 & 0.2242 & 0.2317& \textbf{0.2168}\\

\small{rRMSE$_G^{uint}$} & 0.0587& -& - & 0.0353& \textbf{0.0350}& 0.2804 & -& - &0.1525&\textbf{0.1482} & 0.3026&0.234 & -&- & 0.1796& \textbf{0.1747}\\

\small{RMSE}& -& 1.70& 1.04& 0.6407& \textbf{0.6324} & -& 5.61&2.74&\textbf{2.550}&2.613&12.44& -& 4.44& 2.92& 2.86& \textbf{2.83}\\

\small{RMSE$_G$} & -& 3.24& 1.96& 1.35& \textbf{1.33}& -& 20.13& 6.70&\textbf{5.77}&5.80& 8.06& -& 9.56 & 5.17&5.12&\textbf{4.92}\\

\small{RMSE$_G^{uint}$} & 1.98& -& -& 1.25& \textbf{1.23} &4.76 & -& -&\textbf{3.4924}&3.5275 & 8.06& 5.27& - & -&\textbf{3.66}&\textbf{3.66}\\
\end{tabular}
}
\vspace{-0.3cm}
\end{table*}

\begin{table}[th!]
\caption{NTIRE challenge~\cite{arad2018ntire} results on the test data.}
\centering
\renewcommand\tabcolsep{10pt} 
\resizebox{\linewidth}{!}
{
\begin{tabular}{l||l|r||l|r}
     		 & \multicolumn{2}{c||}{Track 1: Clean} 			& \multicolumn{2}{c}{Track 2: Real World} \\
team 		 & MRAE         		& RMSE        		& MRAE           & RMSE           \\ \hline\hline
VIDAR1~\cite{Shi_2018_CVPR_Workshops}	 & \textbf{0.0137}	        & \textbf{14.45}		& \textbf{0.0310} & \textbf{24.06}          \\
HypedPhoti~\cite{arad2018ntire} 	 & 0.0153        		& 16.07      		& 0.0332          & 27.10          \\
LFB~\cite{Stiebel_2018_CVPR_Workshops} 		 & 0.0152        		& 16.19       		& 0.0335          & 26.44          \\
IVRL Prime~\cite{arad2018ntire}& 0.0155	   			& 16.17 & 0.0358          & 28.23          
\\
sr402~\cite{arad2018ntire}  		 & 0.0164        		& 16.92       		& 0.0345          & 26.97          \\
ours & 0.0174  	   		& 17.27        		& 0.0364          & 27.09          \\
\end{tabular}
}
\label{tab:challenge_results}
\vspace{-0.5cm}
\end{table}

\begin{table*}[th]
\small
\caption{\small{Average RMSE $(\times 10^{-2})$, MRAE and SSIM of spectral reconstructions on ICVL and CAVE images for Continuous and Discrete sensitivity sets. Generic model, Conditional model with estimated sensitivities and Conditional model with groundtruth sensitivities. }}
\label{tab:errors_vs_sensitivity_sets}
\centering
\vspace{0.2cm}
{
\resizebox{\linewidth}{!}{
\begin{tabular}{l|c|c|c|c|c|c|c|c|c|c|c|c}    
&\multicolumn{4}{|c|}{\textbf{RMSE  \textbf{$(\times 10^{-2})$}}}   &\multicolumn{4}{|c|}{\textbf{MRAE}}      &\multicolumn{4}{|c}{\textbf{SSIM}} \\
\hline
    &\multicolumn{2}{|c|}{\textbf{ICVL}}      &\multicolumn{2}{|c}{\textbf{CAVE}}&\multicolumn{2}{|c|}{\textbf{ICVL}}      &\multicolumn{2}{|c}{\textbf{CAVE}}&\multicolumn{2}{|c|}{\textbf{ICVL}}      &\multicolumn{2}{|c}{\textbf{CAVE}} \\
    \hline
    &  Cont & Disc &  Cont & Disc
    &  Cont & Disc &  Cont & Disc
    &  Cont & Disc &  Cont & Disc\\
    \hline
      Generic& $   2.4663   $ &$  2.9379   $ &$    12.0432    $ & $ 12.9653      $& $   0.1050   $ &$  0.1099   $ &$    0.5836    $ & $ 0.6099      $& $   0.9658   $ &$  0.9568   $ &$    0.3510  $ & $  0.2752 $ \\

    \hline
     \textbf{Cond. Est.}& $ 2.0661     $ &$  2.2393    $ &$ 6.3407       $ & $   5.6070    $& $ 0.0855     $ &$ 0.1173    $ &$  0.4083       $ & $   0.3729 $& $ 0.9810     $ &$  0.9754    $ &$ 0.8623 $ & $   0.8683  $
           \\
    \hline
         Cond. GT.& $ 0.8330     $ &$  0.8979   $ &$   3.1699         $ & $   3.6868    $& $  0.0368    $ &$   0.0505  $ &$   0.2893 $ & $  0.3638     $& $  0.9882    $ &$   0.9859  $ &$   0.8915  $ & $  0.8738  $\\

\hline
\end{tabular}
}
}
\vspace{-0.3cm}
\end{table*}

\noindent\textbf{Number of sensitivity functions.}
We demonstrate the behavior of proposed methods for different number of sensitivity functions. By evaluating these controlled cases, we decide on which method to use in the wild. For this purpose, we limit the set of possible sensitivities both in training and testing phases. For each model, limited number of sensitivity functions are selected randomly from our dataset consisting of 40 functions.
Then, each model is trained using those sensitivities. This procedure is repeated five times.

\noindent\textbf{Classifier accuracy.} 
Controlled settings also enables the use of a classifier model.
Figure~\ref{fig:classification} shows the average accuracy and reconstruction error of our classifier model with respect to the number of class labels on ICVL test data.
The output label of the classifier is used to select which specialized model will be used to perform spectral reconstruction.
Figure~\ref{fig:std_graph} demonstrates the spectral reconstruction error depending on the number of sensitivity functions and the model used.
In the ideal case the classifier has perfect accuracy and the correct specialized models is selected for each test image to achieve
the best possible results. However, the accuracy of the classifier decreases as the number of possible labels increases.
Therefore, it can only be used for controlled cases where small set of sensitivity functions present.
On the other hand, generic model and conditional model provide more stable results.

\noindent\textbf{Single camera standard benchmarks. }
To validate our efficient network for the task of example-based spectral reconstruction from an RGB input image,
we adhere to the default settings and train models and report our results on the standard 
ICVL~\cite{arad2016sparse},
CAVE~\cite{yasuma2010generalized}, NUS~\cite{nguyen2014training} and NTIRE~\cite{arad2018ntire} benchmarks.
Note that these benchmarks assume a single camera / sensitivity function to synthesize RGB image from the HS data.
Table~\ref{tab:quantitative_comparison_ICVL_CAVE_NUS} compares our results to the best results reported on ICVL, CAVE and NUS to date.
We report also the results (\textit{ours+E}) after enhancing the prediction by applying our model on 8 images obtained by rotation and flip
and averaging the results after mapping them back.
Our network compares favorable to state-of-the-art, substantially improving the results on ICVL and CAVE.

For the NTIRE 2018 challenge~\cite{arad2018ntire} we adhere to the same settings and deploy a variant of our solution trained for l2-norm loss.
NTIRE has the largest training dataset to date, therefore we add 2 extra blocks to our network and keep all the other settings of the default configuration as used to report results on the ICVL, CAVE, and NUS benchmarks.
Table~\ref{tab:challenge_results} reports our results on NTIRE in comparison with top challenge entries.
For Track 1: Clean conditions, our solution ranks below 5 solutions in MRAE terms. 
For Track 2: Real world, our solution ranks below 3 solutions in RMSE terms.
While still competitive on NTIRE, our solution is the most efficient - lowest number of layers and runtime.  

\noindent\textbf{Unconstrained settings.} 
As a final test, the limits of the generic and conditional models are evaluated.
For this reason, the continuous set is used in training models which spans a large space of functions.
Table~\ref{tab:errors_vs_sensitivity_sets} shows the spectral reconstruction errors of these models.
Although different sensitivity functions are used, the generic model reconstructs the hyperspectral image by
extracting features regarding sensitivity in hidden layers. Since the variations in the sensitivity mostly
show its effect on color balance, extraction of such features is possible from small patches where almost no spatial information exists.
The proposed conditional model performs better on average compared to the generic model. 
The error on CAVE dataset is much larger than the ICVL since the corresponding models are trained with less data.
Applying generic model on CAVE even causes corruptions in several bands depending on the sensitivity used which results in lower overall SSIM scores.
	
\noindent\textbf{Potential of the conditional model.} 
As shown in Fig.~\ref{fig:std_graph} and Tab.~\ref{tab:errors_vs_sensitivity_sets} the performance gap between conditional and generic models is substantial despite the degradation caused by the estimated sensitivity.
The deformation induced on the shape of the responses also affects conditioning since such responses are not included in the training data.
The last case in Tab.~\ref{tab:errors_vs_sensitivity_sets} puts forward the capacity of the proposed method, it shows how much the model can perform if the sensitivity is estimated with perfect accuracy.
The numerical results confirm that a conditional network can perform almost as much as a specialized model. 

\noindent\textbf{Visual Assessment.}
Figure~\ref{fig:ICVL_visual} depicts a visual comparison between the conditional model and the generic model where the RGB image is generated with
a function sampled from our continuous set. The groundtruth of the data is also shown.
Figure~\ref{fig:ICVL_visual_rmse_vs_band} also states the reconstruction error corresponding to each band in the same image.
The results support that our conditional model performs better reconstructions as also stated by the numerical results.

\subsection{Application in the Wild}

\begin{figure}[t]
\centering
\includegraphics[width=1\linewidth]{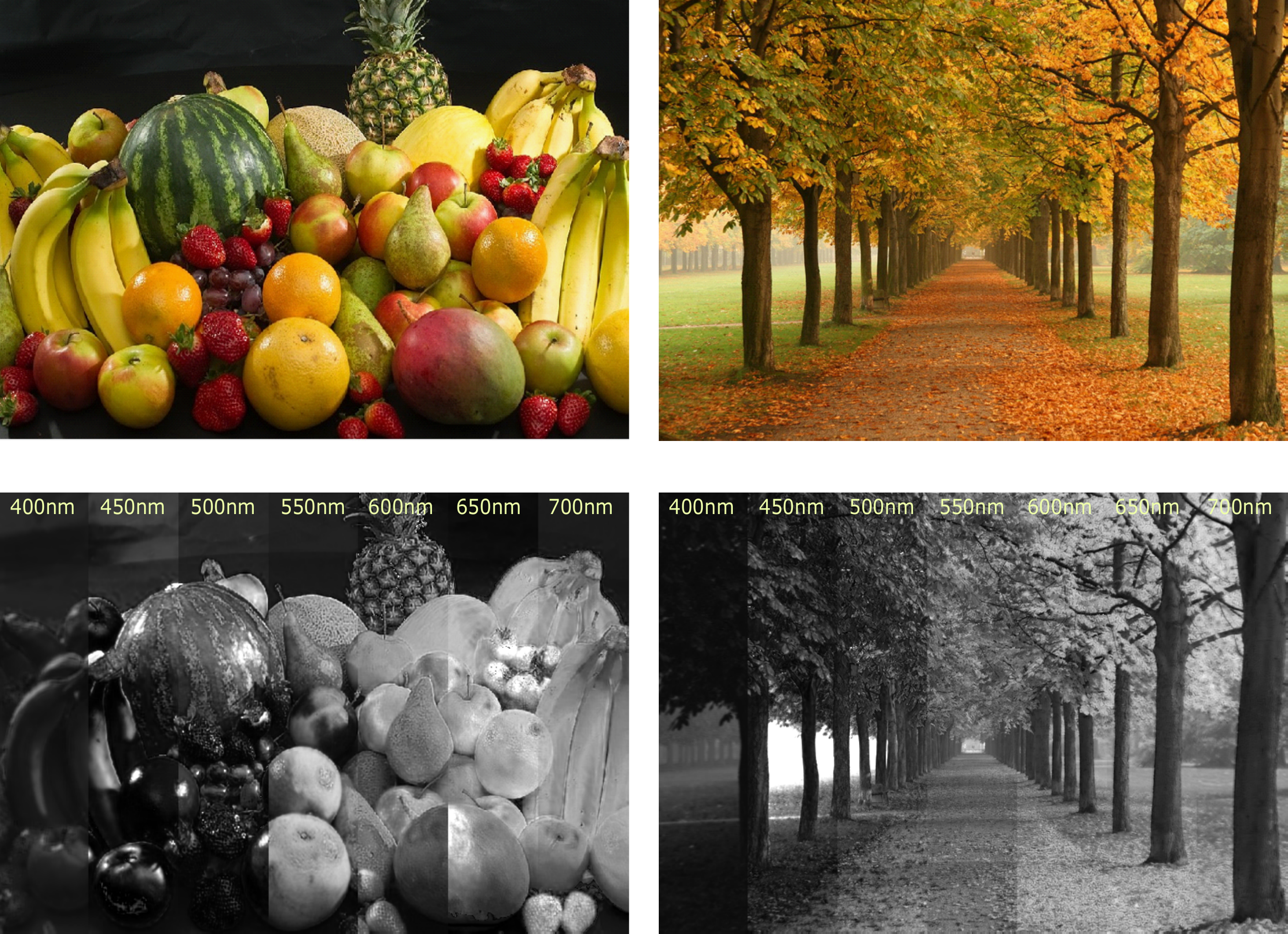}  
\caption{Application of the proposed method on real life examples.
The conditional model is applied on images with unknown settings (top) by estimating the sensitivity function and some of the reconstructed bands are illustrated (bottom). More examples are presented in the suppl. material. }
\label{fig:examples}
\vspace{-0.3cm}
\end{figure}

Our ultimate goal is to estimate the spectral information from an RGB image taken under unknown settings.
In order to exhibit the applicability of our solution, the proposed conditional model is tested on other RGB data.
Figure~\ref{fig:harvard_visual} demonstrates the spectral result estimated with our model trained using ICVL data, its reference spectrum,
along with the RGB image and its corresponding spectral image from the Harvard dataset~\cite{chakrabarti2011statistics}.
Figure~\ref{fig:examples} also presents the reconstructions performed on RGB images with unknown source.
These visuals also verify the feasibility of our approach in real life settings.

\section{Conclusion}
\label{sec:conclusion}
In this paper, we took steps forward towards the estimation of the spectral information from a single RGB image in the wild, with unknown settings.
First, we proposed an efficient neural network for example-based spectral reconstruction from RGB images where there is assumed a known camera and prior availability of training pairs of RGB and spectral images. This approach compares favorable in accuracy and/or efficiency with the current state-of-the-art methods on ICVL, CAVE, NUS and NTIRE standard benchmarks.
Second, we proposed estimator and classifier models to reveal the sensitivity function that would likely fit an RGB image for a reference hyperspectral image.
Third, we combined the sensitivity function estimation with our spectral reconstruction model for different settings.

Our experiments demonstrated that an efficient scheme employing estimation of the sensitivity function and conditioning the spectral reconstruction model is capable of good accuracy for specialized models in the wild.
To best of our knowledge, our work is a first successful attempt to estimate spectral data from a single RGB image captured in unconstrained settings. 
\newpage
{\small
\bibliographystyle{ieee}
\bibliography{egbib}
}
\end{document}